\documentclass[final]{cvpr}

\usepackage{times}
\usepackage{epsfig}
\usepackage{graphicx}
\usepackage{amsmath}
\usepackage{amssymb}

\usepackage{algorithmic, algorithm}
\usepackage{balance}
\usepackage{bm}
\usepackage{booktabs}
\usepackage{caption}
\usepackage{enumitem}
\usepackage{url}
\usepackage{xcolor}
\usepackage{multirow}

\usepackage[pagebackref=true,breaklinks=true,colorlinks,bookmarks=false]{hyperref}

\def\confYear{CVPR 2021}

\begin{document}

\title{Object-Centric Neural Scene Rendering}

\author{Michelle Guo\\
Stanford University
\and
Alireza Fathi\\
Google Research
\and
Jiajun Wu\\
Stanford University
\and
Thomas Funkhouser\\
Google Research
}


\newcommand{\xpar}[1]{\noindent\textbf{#1}\ \ }
\newcommand{\vpar}[1]{\vspace{3mm}\noindent\textbf{#1}\ \ }

\newcommand{\sect}[1]{Section~\ref{#1}}
\newcommand{\ssect}[1]{\S~\ref{#1}}
\newcommand{\eqn}[1]{Equation~\ref{#1}}
\newcommand{\fig}[1]{Figure~\ref{#1}}
\newcommand{\tbl}[1]{Table~\ref{#1}}
\newcommand{\alg}[1]{Algorithm~\ref{#1}}

\newcommand{\shapenet}{ShapeNet\xspace}
\newcommand{\pascal}{PASCAL 3D+\xspace}
\newcommand{\suncg}{SUNCG\xspace}
\newcommand{\scenenet}{SceneNet RGB-D\xspace}
\newcommand{\sunrgbd}{SUN RGB-D\xspace}

\newcommand{\degree}{\ensuremath{^\circ}\xspace}
\newcommand{\ignore}[1]{}
\newcommand{\norm}[1]{\lVert#1\rVert}
\newcommand{\fcseven}{$\mbox{fc}_7$}

\renewcommand*{\thefootnote}{\fnsymbol{footnote}}

\def\naive{na\"{\i}ve\xspace}
\def\Naive{Na\"{\i}ve\xspace}

\makeatletter
\DeclareRobustCommand\onedot{\futurelet\@let@token\@onedot}
\def\@onedot{\ifx\@let@token.\else.\null\fi\xspace}

\def\eg{\emph{e.g}\onedot} \def\Eg{\emph{E.g}\onedot}
\def\ie{\emph{i.e}\onedot} \def\Ie{\emph{I.e}\onedot}
\def\cf{\emph{c.f}\onedot} \def\Cf{\emph{C.f}\onedot}
\def\etc{\emph{etc}\onedot} \def\vs{\emph{vs}\onedot}
\def\wrt{w.r.t\onedot} \def\dof{d.o.f\onedot}
\def\etal{\emph{et al}\onedot}
\makeatother

\definecolor{MyDarkBlue}{rgb}{0,0.08,1}
\definecolor{MyDarkGreen}{rgb}{0.02,0.6,0.02}
\definecolor{MyDarkRed}{rgb}{0.8,0.02,0.02}
\definecolor{MyDarkOrange}{rgb}{0.40,0.2,0.02}
\definecolor{MyPurple}{RGB}{111,0,255}
\definecolor{MyRed}{rgb}{1.0,0.0,0.0}
\definecolor{MyGold}{rgb}{0.75,0.6,0.12}
\definecolor{MyDarkgray}{rgb}{0.66, 0.66, 0.66}

\def\bR{\mathbb{R}}
\def\cL{\mathcal{L}}
\def\vr{\bm{r}}
\def\vx{\bm{x}}
\def\vw{\bm{\omega}}
\def\outdir{\bm{\omega_o}}
\def\indir{\bm{\omega_l}}
\def\transform{\bm{T}}
\def\dim{\bm{D}}

\def\psnr{\text{PSNR}$\uparrow$}
\def\ssim{\text{SSIM}$\uparrow$}
\def\lpips{\text{LPIPS}$\downarrow$}

\def\model{\mbox{OSF}\xspace}
\def\models{\mbox{OSFs}\xspace}

\newcommand{\myparagraph}{\vspace{-10pt}\paragraph}
\newcommand{\myitem}{\vspace{-3pt}\item}

\twocolumn[{%
\renewcommand\twocolumn[1][]{#1}%
\maketitle
\begin{center}
    \centering
    \vspace{-4mm}
    \includegraphics[width=\linewidth]{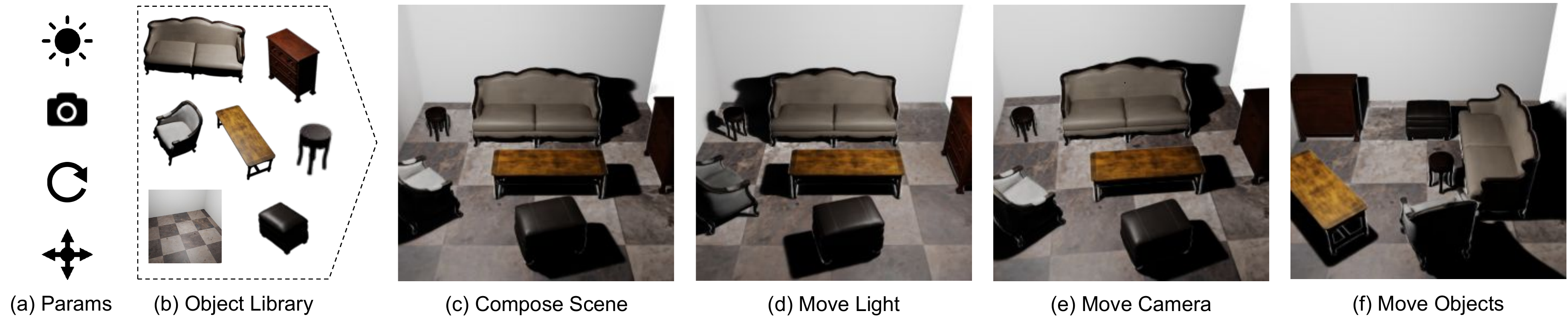}
    \captionof{figure}{We propose an object-centric neural scene representation for image synthesis. Given a scene description (a), and a repository of neural object-centric scattering functions (OSF) trained independently and frozen for each object (b), we can compose the objects into scenes (c), and render photorealistic images as we move lights (d), cameras (e), and/or objects (f).}
    \label{fig:teaser}
\end{center}%
}]

\begin{abstract}
We present a method for composing photorealistic scenes from captured images of objects.
Our work builds upon neural radiance fields (NeRFs), which implicitly model the volumetric density and directionally-emitted radiance of a scene.
While NeRFs synthesize realistic pictures, they only model static scenes and are closely tied to specific imaging conditions.
This property makes NeRFs hard to generalize to new scenarios, including new lighting or new arrangements of objects.
Instead of learning a scene radiance field as a NeRF does, we propose to learn object-centric neural scattering functions (\models), a representation that models per-object light transport implicitly using a lighting- and view-dependent neural network.
This enables rendering scenes even when objects or lights move, without retraining.
Combined with a volumetric path tracing procedure, our framework is capable of rendering both intra- and inter-object light transport effects including occlusions, specularities, shadows, and indirect illumination.
We evaluate our approach on scene composition and show that it generalizes to novel illumination conditions, producing photorealistic, physically accurate renderings of multi-object scenes.
\end{abstract}

\section{Introduction}
Synthesizing images of dynamic scenes is an important problem in computer vision and graphics, with applications in AR/VR and robotics~\cite{habitat19iccv,xia2020interactive}.
For synthetic scenes, a user typically designs a set of 3D objects separately~\cite{blender}, then composes them into scenes to be rendered with specified camera, material, and lighting parameters.   While this traditional graphics approach allows flexible scene compositions, it produces photorealistic images only when detailed models are available for geometry, lighting, materials, and cameras, which are difficult to obtain for real-world scenes.

\begin{figure}
\centering
\includegraphics[width=\linewidth]{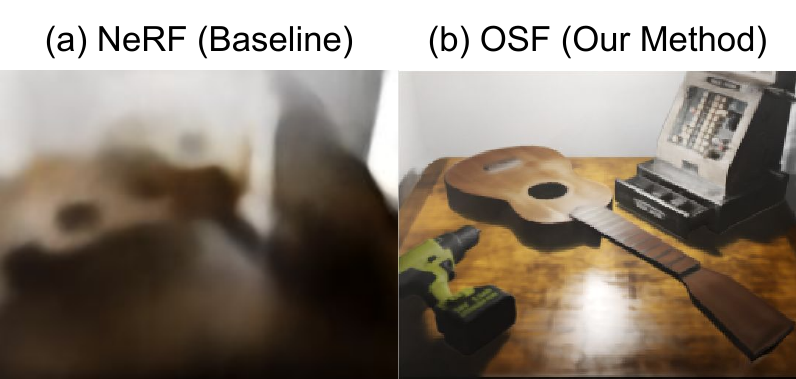}
\caption{(a) Result of learning NeRF on a dynamic scene.
NeRF assumes static scenes and fixed lighting, producing blurred predictions when trained on dynamic scenes where objects and lighting are randomly moved.
(b) In contrast, our method is able to predict crisp images.
}
\label{fig:nerf_blur}
\end{figure}

Recently, a number of works have demonstrated high-quality results on novel view synthesis from real-world images using implicit methods~\cite{lombardi2019neural,sitzmann2019deepvoxels,sitzmann2019scene}.
Most notably, neural radiance fields (NeRFs)~\cite{mildenhall2020nerf} achieve photorealistic quality by implicitly modeling the volumetric density and directional emitted radiance of a scene.
However, a NeRF assumes static scenes and fixed illumination (\fig{fig:nerf_blur}).
It learns a radiance field, which estimates only the resulting radiance along a ray after all light transport has occurred in a scene. 
Thus, for dynamic scenes where lights and objects can move, a NeRF model would have to be trained separately for every new scene configuration.  

To address this issue, we propose object-centric neural scattering functions (\models) to synthesize dynamic scenes of objects learned from 2D images (\fig{fig:teaser}).
We represent each object as a learned 7D scattering function with inputs $(x, y, z, \phi_i,\theta_i,\phi_o,\theta_o)$, where $(x, y, z)$ is the spatial location, $(\phi_i,\theta_i)$ is the incoming light direction, and $(\phi_o,\theta_o)$ is the outgoing light direction.
The function outputs the volumetric density as well as the fraction of light arriving from direction $(\phi_i,\theta_i)$ that scatters in outgoing direction $(\phi_o,\theta_o)$.

Since each object's scattering function is a radiance transfer function rather than a radiance field, it is intrinsic to the object (independent of the scene it is in) and can be reused across different object placements and lighting conditions without retraining.
This allows us to build a library of \models trained independently for different objects that can be composed into scenes with different object placements.

To model light transport \textit{between} objects, we integrate our implicit object functions with volumetric path tracing.  Like NeRF, we evaluate the radiance and volumetric density at 5D samples along every primary ray to the camera and composite them with an over operator.
However, unlike NeRF, we estimate the radiance for each 5D sample by integrating our 7D \model across the 2D sphere of incoming light directions.
We estimate the integral with Monte Carlo path tracing~\cite{kajiya1986rendering} to reproduce shadows and indirect illumination effects.

Our key idea is to decompose the rendering problem into (i) a learned component (per-object ``asset creation''), and (ii) a non-learned component (per-scene path tracing).
The learned component models intra-object light transport (\eg, bounces from the seat of a chair to the back of the chair), the non-learned component handles inter-object light transport (\eg, bounces from a wall to a chair), and 
together they model the full rendering equation~\cite{kajiya1986rendering}, except for occlusions of intra-object illumination by other objects.
Since only the inter-object light transport changes as objects and lights move, no re-training is required for different scene arrangements.
Experimental results indicate that our method is capable of rendering images with 
novel scene compositions and lighting conditions better than alternative learned approaches. In summary, our contributions are
\begin{enumerate}
    \myitem Learning object-centric neural scattering functions (\model) that model intra-object light transport implicitly using a lighting- and view-dependent neural network.
    \myitem Integrating implicitly learned object scattering functions with volumetric path tracing to model inter-object light transport.
    \myitem A rendering algorithm that enables rendering scenes with moving objects, lights and cameras, using implicit functions.
\end{enumerate}

\section{Related Work}
\paragraph{Classical object-centric representations.}
Factoring light transport into intra- and inter-object illumination has a long history in traditional computer graphics~\cite{dutre2018advanced}.
In most cases, the motivation is to improve rendering efficiency by approximating intra-object lighting factors with linear transformations, as in precomputed radiance transfer (PRT)~\cite{sloan2002precomputed}, ambient occlusion~\cite{miller1994efficient}, or virtual walls~\cite{arnaldi1994division}.
In other cases, the motivation is to insert captured, real-world radiance fields into synthetic scenes, as in Light Field Transfer~\cite{cossairt2008light}.
These methods generally store the radiance field for objects in a discrete representation (\eg, a sampled 2D or 4D grid).
As a result, they cannot reproduce accurate inter-object light transport, especially for objects with intersecting bounding volumes.
In contrast, we encode the \emph{scattering} field for an object in a 7D continuous representation encoded in the weights of a deep network and utilize volumetric rendering techniques to account for inter-object illumination.

\myparagraph{Novel view synthesis.}
Traditional methods for synthesizing novel views of a scene from captured images include using Structure-From-Motion~\cite{hartley2003multiple} and bundle adjustment~\cite{triggs1999bundle} to predict a sparse point cloud and camera parameters of the scene.
More recently, a number of learning-based novel view synthesis methods have been presented but require 3D geometry as inputs~\cite{hedman2018deep,thies2019deferred,meshry2019neural,aliev2019neural,martin2018lookingood}.
Others use multiplane images as proxies for novel view synthesis, but their viewing ranges are limited to interpolated input views~\cite{flynn2016deepstereo,zhou2018stereo,srinivasan2019pushing,mildenhall2019local}.
Some works represent scenes as coarse voxel grids and use a CNN-based decoder for differentiable rendering, but lack view consistency due to the use of 2D convolutional kernels~\cite{nguyen2018rendernet,nguyen2019hologan,nguyen2020blockgan}.

Recently, volume rendering approaches have been used to render scenes represented as voxel grids that are more view-consistent~\cite{lombardi2019neural,sitzmann2019deepvoxels}.
However, the rendering resolution of these methods are limited by the time and computational complexity of discretely sampled volumes.
To address this issue, Neural Radiance Fields (NeRF)~\cite{mildenhall2020nerf} directly optimizes a \textit{continuous} radiance field representation using a multi-layer perceptron.
This allows synthesizing novel views of realistic images at an unprecedented level of fidelity.
To make NeRF more efficient, Neural Sparse Voxel Fields~\cite{liu2020neural} proposes a sparse voxel octree variant of NeRF, and demonstrates the ease of composing learned NeRFs with their voxel representation.
While these implicit methods produce high-quality novel views of a scene, their models assume a static scene with fixed illumination.
Our method enables synthesizing dynamic scenes with novel viewpoint, lighting, and object configurations.

\myparagraph{Relighting.}
Learning-based methods that relight images without explicit geometric reasoning have been proposed, but lack the ability to recover hard shadows~\cite{sun2019single,xu2018deep,zhou2019deep}.
Other works use geometric representations that facilitate shadowing computation, but require 3D geometry as input~\cite{philip2019multi,zhang2020neural,oechsle2020learning,rematas2020neural}.
Deep Reflectance Volumes~\cite{bi2020deep} reconstructs a voxelized representation of a scene and predict per-voxel BRDFs, but the fixed resolution of voxel grids limits the quality in the rendered images.
Similarly, Neural Reflectance Fields~\cite{bi2020neural} predicts the parameters of a BRDF model, but demonstrate higher fidelity rendering by learning a continuous scene representation.
However, Neural Reflectance Fields focuses on relighting single objects or scenes, and requires manual specification of the BRDF model.
In contrast, our method does not assume knowledge of the material model of captured objects, and can render \textit{multiple} objects in dynamic scenes.

\section{Preliminaries}
\label{sec:prelim}

In this section, we provide background on existing methods for volume rendering~\cite{novak2018monte} and NeRF~\cite{mildenhall2020nerf}, which are prerequisites for our methods (\sect{sec:method}).
We first discuss the volume rendering equation underlying our approach (\sect{sec:volume_rendering}).
Then we introduce a numerical approximation to compute the equation using ray marching (\sect{sec:ray_marching}).
Finally, we provide background on NeRF (\sect{sec:nerf}), a method that uses volume rendering and a continuous scene representation for rendering scenes.

\subsection{Volume Rendering}
\label{sec:volume_rendering}
To render an image of a scene with arbitrary camera parameters, camera rays are sent into the scene, through each pixel on the image plane.
The expected color of each pixel is computed as the radiance along each camera ray.

Volume rendering is an approach for computing the radiance traveling along rays traced in a volume.
Let $\vr(t)=\vx_0+\outdir t$ be a point along a ray $\vr$ with origin $\vx_0$ and direction $\outdir$, where $t\in\mathbb{R}$ is a 1D location along the ray, and the $\bm{o}$ in $\outdir$ denotes ``outgoing'' direction.
For our purposes, we assume non-emissive and non-absorptive volumes.
The volume rendering equation~\cite{novak2018monte} to compute the radiance $L(\vx_0, \outdir)$ of the ray is defined as the following:
\begin{equation}
\label{eqn:volume_rendering}
    L(\vx_0, \outdir) = \int_{t_n}^{t_f}\tau(t)\sigma(\vr(t))L_s(\vr(t),\outdir)dt\text{,}
\end{equation}
where $t_n$ and $t_f$ are near and far integration bounds and $\sigma(\vr(t))$ denotes the volume density of point $\vr(t)$.
The $\tau(t)$ is computed as $\exp{\left(-\int_{t_n}^t \sigma(\vr(u))du\right)}$, which denotes the accumulated transmittance from $t_n$ to $t$.
$L_s\left(\vr(t),\outdir\right)$ is the amount of light scattered at point $\bm{r}(t)$ along direction $\outdir$, defined as the integral over all incoming light directions:
\begin{equation}
    L_s(\vx,\outdir) = \int_{\mathcal{S}}L(\vx,\indir)f_p(\vx,\indir, \outdir)d\indir,
\label{eqn:scatter}
\end{equation}
where $\mathcal{S}$ is a unit sphere and $f_p$ is a phase function that evaluates the fraction of light incoming from direction $\indir$ at a point $\vx$ that scatters out in direction $\outdir$.
Note that NeRF~\cite{mildenhall2020nerf} assumes fixed illumination and does not consider any form of \eqn{eqn:scatter}.
We consider a more general form of the volume rendering equation that explicitly models light paths within (intra) and between (inter) objects in a scene.
This is important for dynamic scenes, where lighting and objects can move with respect to one another.

\subsection{Ray Marching}
\label{sec:ray_marching}
The continuous integrals in \eqn{eqn:volume_rendering} can be numerically estimated with quadrature~\cite{kniss2003model,max1995optical}, as done in NeRF~\cite{mildenhall2020nerf}.
For each ray, stratified sampling is used to obtain $N$ samples $\{t_i\}_{i=1}^N$ along the ray, where $t_i\in[t_n, t_f]$.
The rendering equation can be approximated by:
\begin{equation}
    L(\vx_0, \outdir) = \sum_{i=1}^{N}\tau_i\alpha_i L_s(\bm{x}_i,\outdir),
    \label{eqn:ray_marching}
\end{equation}
where $\tau_i=\prod_{j=1}^{i-1}(1-\alpha_j)$ and $\alpha_i=1-e^{-\sigma_i(t_{i+1}-t_i)}$.
To compute $L_s$, we discretize over the domain $\mathcal{S}$ in \eqn{eqn:scatter} by sampling a set of incoming light paths $\mathcal{L}=\{\bm{l}_1, \ldots, \bm{l}_K\}$.
We can compute $L_s$ as the average over the light paths:
\begin{equation}
    L_s(\vx_i,\outdir) = \frac{1}{\lvert\mathcal{L} \rvert}\sum_{l\in\mathcal{L}}L(\vx_i,\indir)\bm{\rho^l}_i,
\end{equation}
where $\bm{\rho^l}_i=f_p(\vx_i,\indir, \outdir)\in [0,1]$, the fraction of light incoming from light path $\bm{l}$ that is scattered in direction $\outdir$.

\subsection{Neural Radiance Fields}
\label{sec:nerf}
NeRF represents a continuous scene as a volumetric radiance field, approximated with a multilayer perceptron $F_\Theta$.
The model $F_\Theta$ takes spatial location $\bm{x}=(x, y, z)$ and viewing direction $\bm{d}=(\phi,\theta)$ as input,
and outputs the density $\sigma$ and color $\bm{c}=(r, g, b)$, where $r, g, b \in[0,1]$.
Frequency-based positional encoding~\cite{rahaman2019spectral,vaswani2017attention} is applied to the inputs to better capture high-frequency variation in appearance and geometry.

A hierarchical volume sampling procedure~\cite{mildenhall2020nerf,levoy1990efficient} is then employed to more efficiently allocate samples along each ray.
This technique biases sample allocation to favor the visible parts of the scene that contribute the most to the final render, avoiding occluded or free space in the scene.
NeRF simultaneously optimizes two radiance fields, where the sample weights $\tau_i\cdot\alpha_i$ from a \textit{coarse} model are used to bias samples for a \textit{fine} model.
The $L_2$ loss is used to optimize both models:
\begin{equation}
    \sum_{\bm{r}\in\mathcal{R}} \norm{\widehat{C}_c(\bm{r}) - C(\bm{r})}^2_2 + \norm{\widehat{C}_f(\bm{r}) - C(\bm{r})}^2_2,
\end{equation}
where $\mathcal{R}$ is the set of all camera rays, $\widehat{C}_c(\vr)$ and $\widehat{C}_f(\vr)$ denote the radiance along $\vr$ predicted by the coarse and fine models respectively, and $C(\vr)$ is the ground truth pixel color corresponding to ray $\bm{r}$.

\section{Method}
\label{sec:method}

\subsection{Object-Centric Neural Scattering Function}
\begin{figure}
\centering
\includegraphics[width=\linewidth]{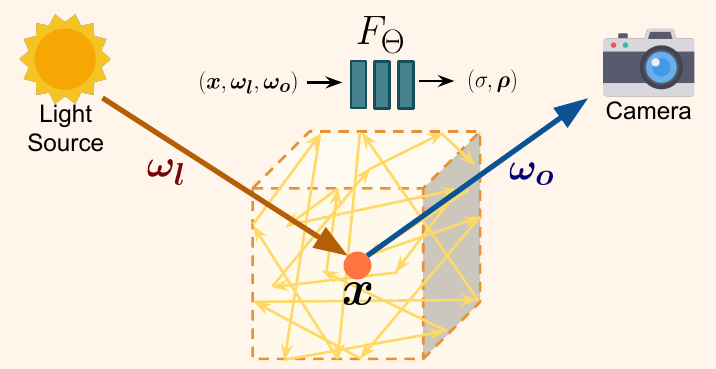}
\caption{We represent each object as an object-centric neural scattering function (\model), which models how light entering at a point $\vx$ on the object, from direction $\indir$ where $\bm{l}$ corresponds to a light path, undergoes multiple bounces within the object and exits along direction $\outdir$ with some fractional amount of light $\bm{\rho}$.
We approximate the scattering function with a multilayer perceptron $F_\Theta$ where $\Theta$ are learned weights that parameterize the neural network.
Given a single point $\vx$, an incoming light direction $\indir$, and an outgoing direction $\outdir$, $F_\Theta$ outputs the volume density $\sigma$ of that point, as well as the fraction of light arriving at $\vx$ from direction $\indir$ that is scattered in direction $\outdir$.
}
\label{fig:methods_single}
\end{figure}

\label{sec:onerfs}
We represent each object as a 7D object-centric neural scattering function (\model), depicted in \fig{fig:methods_single}.
For each object, we learn an implicit function $F_\Theta \colon (\vx, \indir, \outdir) \rightarrow (\sigma, \bm{\rho})$ that receives a 3D point in the object coordinate frame, the incoming light direction, and the outgoing light direction, and predicts the volumetric density as well as fraction of incoming light that is scattered in the outgoing direction.
$\Theta$ are learned weights that parameterize the neural network, $\vx=(x, y, z)$ denotes the spatial location, $\indir=(\phi_l,\theta_l)$ denotes the incoming light direction,
$\outdir=(\phi_o, \theta_o)$ denotes the outgoing light direction, $\sigma$ denotes the volumetric density, and $\bm{\rho}=(\rho_r, \rho_g, \rho_b)$ denotes the fraction of light arriving at $\vx$ from direction $\indir$ that is scattered and leaving in direction $\outdir$.
The final color of a point $\vx$ is the integral of $\bm{\rho}$ multiplied by the incoming radiance over all incoming light directions in unit sphere $\mathcal{S}$ (\eqn{eqn:scatter}).
Following NeRF, we similarly apply positional encoding to our inputs $(\vx, \indir, \outdir)$ and employ a hierarchical sampling procedure to recover higher quality appearance and geometry of learned objects.

During training, we assume a single point light source with radiance of $(1, 1, 1)$.
This simplifies $L_s$ from \eqn{eqn:scatter} to the following:
\begin{equation}
\small
    L_s(\vx,\outdir) = L(\vx,\indir)f_p(\vx,\indir, \outdir) = f_p(\vx,\indir, \outdir).
\end{equation}

\noindent To learn per-object NeRFs independent of object rotation and translation, the inputs to $F_\Theta$ must be in the object's canonical coordinate frame.
Given a object transformation $T_i$ for object $\bm{o}_i$, we apply $T_i^{-1}$ to $(\vr,\indir,\outdir)$ before feeding the inputs to the network.

\subsection{Rendering Multiple \models}
\label{sec:rendering_onerfs}
\begin{figure}
\centering
\includegraphics[width=\linewidth]{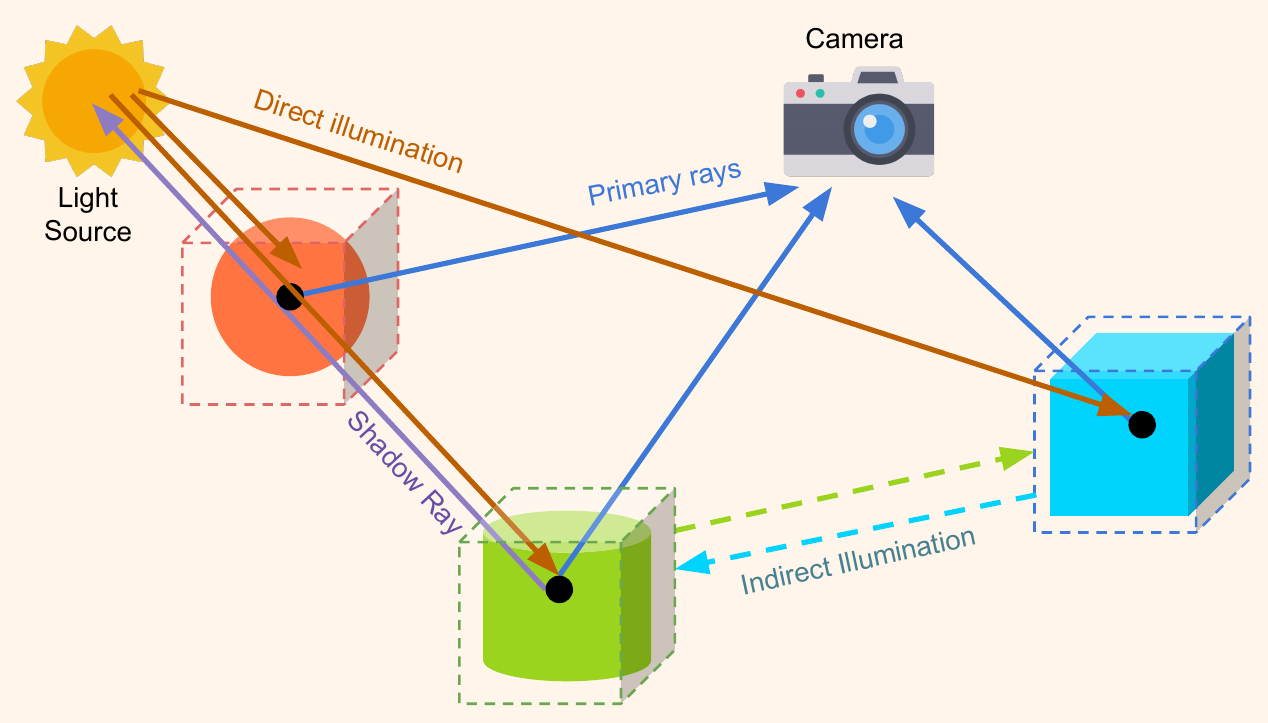}
\caption{Rendering multiple object-centric neural scattering functions (\models).
We propose a rendering procedure for rendering an arbitrary scene description consisting of objects, light sources, and camera information.
Given a set of objects, we compute direct illumination by shooting rays from each light source to each object (brown arrows).
Shadows are computed by sending shadow rays
back to each light source (purple arrow).
In this scenario, the shadow ray from the green cylinder is occluded by the red ball, so the red ball casts a shadow on the green cylinder.
We also send secondary rays between objects to render indirect illumination effects, such as between the green and blue objects in the illustration (green and blue dashed arrows).
Finally, rays are sent back to the camera to render the final image (dark blue arrows).
}
\label{fig:methods_multi}
\end{figure}

Once we have learned an \model for each object, we aim at composing the learned objects into scenes.
An overview of our procedure is visually depicted in \fig{fig:methods_multi}.

\begin{figure}
\centering
\includegraphics[width=\linewidth]{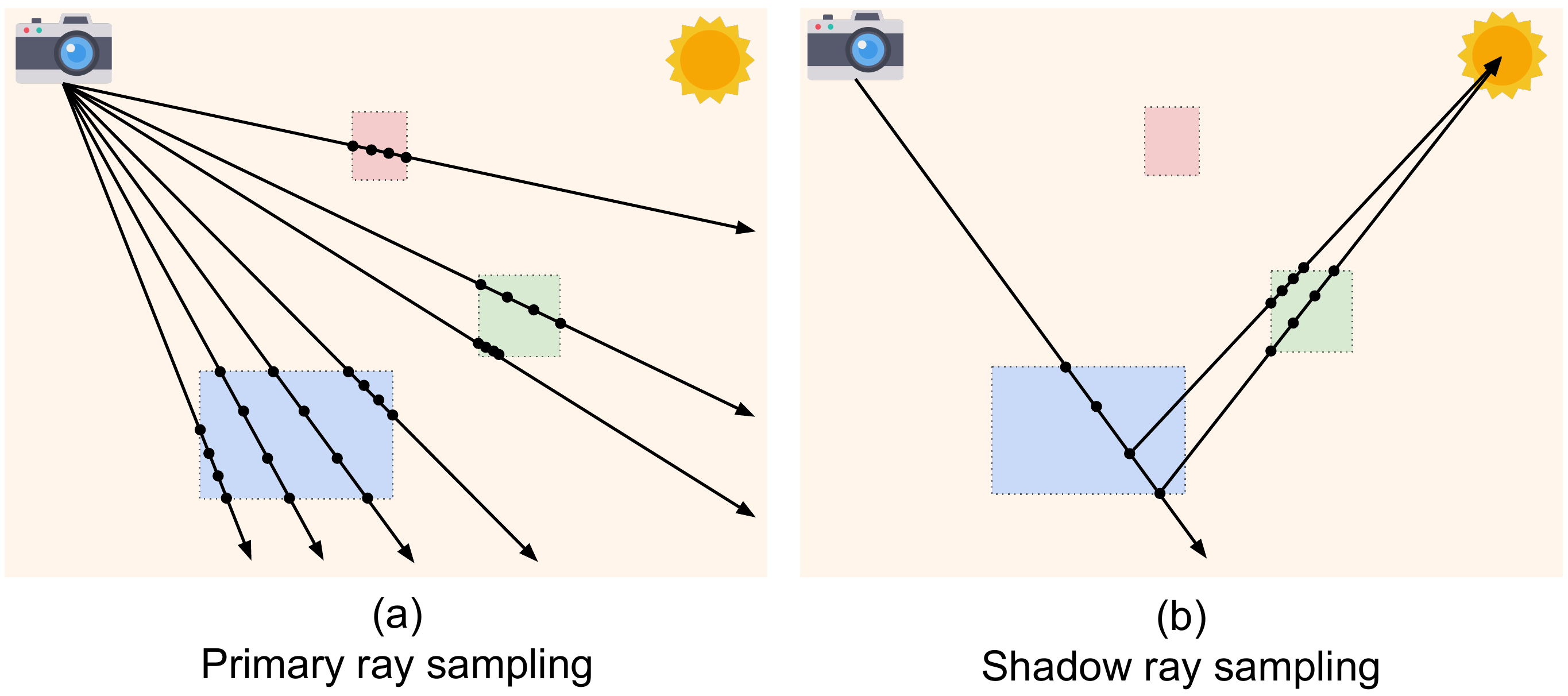}
\caption{\model sampling procedure. Given a scene with a camera, light source, and object bounding boxes, we first send camera rays from the camera into the scene (a). Rays that do not intersect with objects are pruned. Of the rays that do intersect, we compute ray-box intersections and sample points within intersecting regions. To compute shadows, we send shadow rays from each sample to the light source, and evaluate samples within intersecting regions along the shadow rays (b). A similar sampling procedure is used for evaluating secondary rays for indirect illumination.}
\label{fig:methods_sampling}
\end{figure}

Let $\mathcal{O}=\{\bm{o_i}\}_{i=1}^{N}$ be a set of $N$ objects we wish to render.
For simplicity, we first describe the rendering process for each object $\bm{o}_i$, then explain the process to combine results across all objects to render the final scene.
Let $\bm{o}_i\in\mathcal{O}$ denote object $i$ with transformation $T_i\in\mathbb{R}^{4\times4}$ and bounding box dimensions $D_i\in\mathbb{R}^3$.
Further let $\vr$ be a camera ray with origin $\bm{c}\in\mathbb{R}^3$ and direction $\outdir\in\mathbb{R}^3$, which we define with parameters $\gamma=[\bm{c},\outdir]\in\mathbb{R}^6$.
Our goal is to compute $L(\bm{c}, \outdir)$ as described in \eqn{eqn:ray_marching}.
We compute the ray-box intersection between the ray and the object to obtain near bound $t^i_n$ and far bound $t^i_f$ such that $\vr(t^i_n)$ and $\vr(t^i_f)$ each intersect a box plane, as shown in \fig{fig:methods_sampling}.
Note that rays that do not intersect with $\bm{o}_i$ are excluded from our computation.
We sample $M$ points between $t^i_n$ and $t^i_f$ along ray $\vr$ to obtain a sample  $\bm{X}^i=\{\bm{x}^i_m\}_{m=1}^M$, where $\bm{X}^i\in\mathbb{R}^{M\times3}$.
Given a light source $\bm{l}$, we evaluate the object's model $F_{\Theta_i}(\bm{X}^i, \indir, \outdir)$ to obtain alpha values $\bm{\alpha}^i\in\mathbb{R}^{M}$ and phase function values $\bm{\rho}^i\in\mathbb{R}^{M\times3}$. 

It is not always possible for a light ray from light source $\bm{l}$ to reach the object $\bm{o}_i$.
Any of the other objects in $\mathcal{O}'=\{\bm{o}_j\in\mathcal{O}\mid j\neq i\}$ in the scene may occlude the incoming light, casting a shadow on object $\bm{o}_i$.
We compute shadows by sending a shadow ray $\vr_m$ from each of the $M$ samples in $\bm{X}^i$ to the light source $\bm{l}$.
Evaluating the shadow ray enables us to determine the amount of light blocked along the ray by other objects.
We define the parameters of the $M$ shadow rays as $\Gamma\in\mathbb{R}^{M\times 6}$.

For each object $\bm{o}_j\in\mathcal{O}'$, we compute ray-box intersections between shadow rays $\Gamma$ and $\bm{o}_j$'s bounding box.
This allows us to compute the amount of light traveling towards $\bm{o}_i$ that is blocked by $\bm{o}_j$.
Similar to primary rays, we sample $M$ points along each shadow ray to obtain a set of points $\bm{X}^j\in \mathbb{R}^{M\times M}$.
We then evaluate the object model $F_{\Theta_j}(\bm{X}^{j})$ to obtain alpha values $\bm{A}^j\in \mathbb{R}^{M\times M}$.
For each shadow ray $\vr_m$, we combine samples $\bm{A}^j_m$ across the $N-1$ objects in $\mathcal{O}'$ by sorting according to sample distance to obtain alpha values $\bm{A}_m\in\mathbb{R}^{M(N-1)}$.
The fraction of unobstructed light traveling along the shadow ray $\vr_m$ is computed as the transmittance:
\begin{equation}
    \tau_m^l=\prod_{n=1}^{M(N-1)}\left(1 - \bm{A}_{mn}\right).
    \label{eq:tau_m}
\end{equation}
Thus, the adjusted incoming radiance from light source $\bm{l}$ when accounting for occlusions is computed as $L_{\bm{l}}(\bm{x}_m,\indir)=\tau_m^{\bm{l}}L_{\bm{l}}(\bm{x}_m,\indir)$.

We follow the scattering equation in \eqn{eqn:scatter} and now consider all incoming light directions over the unit sphere $\mathcal{S}$.
This accounts for secondary light rays traveling to an object $\bm{o}_i$ indirectly from another object $\bm{o}_j$ (indirect illumination).
We approximate the integral over the unit sphere $\mathcal{S}$ by sampling $K$ directions on the unit sphere uniformly at random.
For each direction $\bm{\omega_k}$ randomly sampled for a point $\vx$, we send a secondary ray $\vr_k$ from $\vx$ in direction $\bm{\omega_k}$ and evaluate the radiance $L(\vx, \bm{\omega_k})$ traveling along the ray.
To compute the radiance of the secondary ray $L(\vx, \bm{\omega_k})$, we employ the same technique used to compute the radiance of a primary ray $L(\bm{c}, \outdir)$ (described at the beginning of \sect{sec:rendering_onerfs}).
The incoming radiance $L(\vx, \bm{\omega_k})$ is multiplied with the phase function value $\bm{\rho}=f_p(\vx, \bm{\omega_k}, \outdir)$ to determine the outgoing radiance $L(\vx, \outdir)$, where $\bm{\rho}$ is evaluated using $F_{\Theta_i}$.
Note that this is possible due to the recursive nature of our formulation.
Only secondary rays are described here (two bounces), but our method supports an arbitrary number of bounces.

\vpar{Rendering.} We sample and evaluate all objects in $\mathcal{O}$ to obtain alpha values $\{\bm{\alpha}^i\}_{i=1}^N$ and phase function values $\{\bm{\rho}^i\}_{i=1}^N$
for a set of sampled points $\{\bm{X}^i\}_{i=1}^N$ along ray $\vr$.
We sort the samples across all objects to produce a final set of combined $P=M\cdot N$ samples $\{\vx_m\}_{m=1}^{P}$, $\{\alpha_m\}_{m=1}^{P}$, and $\{\bm{\rho}_m\}_{m=1}^{P}$.

Given light paths $\mathcal{L}$ containing both direct and indirect illumination, we render the final radiance of a ray with origin $\vx_0$ and direction $\outdir$ with the following equation:
\begin{equation}
    L(\vx_0, \outdir) = \frac{1}{|\mathcal{L}|}\sum_{\bm{l}\in\mathcal{L}}\sum_{m=1}^{P} \alpha_m\bm{\rho^l}_m\tau_mL_{\bm{l}}(\bm{x}_m,\indir)\text{,} 
    \label{eq:L_o}
\end{equation}
where $\tau_m=\prod_{n=1}^{m-1}(1 - \alpha_n)$ and $L_{\bm{l}}(\vx_m,\indir)$ is the radiance from light path $\bm{l}$  arriving at point $\vx_m$.
Note that $P$ is an upper bound on number of samples that need to be evaluated.
In practice, a single ray often only intersects with at most one object in the scene, which means that the proposed rendering procedure is not significantly more expensive than the single object setting.

\begin{table}[t]
\centering
\begin{tabular}{lccc}
\toprule
 \multirow{2}{*}{Methods} & \multicolumn{3}{c}{Furniture-Single} \\ \cmidrule{2-4}
       & \psnr       & \ssim       & \lpips     \\ \midrule
o-NeRF~\cite{mildenhall2020nerf} & 33.22 & 0.980 & 0.021 \\
\model (ours) & \textbf{44.07} & \textbf{0.998} & \textbf{0.002} \\
\bottomrule
\end{tabular}
\caption{Results on rendering objects with novel illumination.
Rows denote different methods: \model and a variant of NeRF~\cite{mildenhall2020nerf} where one NeRF is trained per object (o-NeRF).
}
\label{tbl:single}
\end{table}

\subsection{Implementation Details}

We approximate our model $F_{\Theta}$ with a multilayer perception (MLP) with rectified linear activations.
The predicted density $\sigma$ is view-invariant, while the scattering function value $\bm{\rho}$ is dependent on the incoming and outgoing light directions.
We use an eight-layer MLP with 256 channels to predict $\sigma$, and a four-layer MLP with 128 channels to predict $\bm{\rho}$.
For positional encoding, we use $W=10$ to encode the position $\bm{x}$ and $W=4$ to encode the incoming and outgoing directions $(\indir,\outdir)$, where $W$ is the highest frequency level.
To avoid $\bm{\rho}$ from saturating in training, we adopt a scaled sigmoid~\cite{brock2016generative} defined as $S'(\bm{\rho}) = \delta(S(\bm{\rho}) - 0.5) + 0.5$ with $\delta=1.2$.
We use a batch size of 4,096 rays, and sample $N_c=64$ coarse samples and $N_f=128$ fine samples per ray.
We use the Adam optimizer~\cite{kingma2014adam} with a learning rate of 0.001, $\beta_1=0.9$, $\beta_2=0.999$, and $\epsilon=10^{-7}$.

\section{Experiments}

\paragraph{Datasets and Evaluation Metrics.}
We create three image datasets for our experimental evaluation:
\begin{enumerate}
    \myitem Furniture-Single: Contains fifteen objects rendered individually on a white background with random object pose, point light location, and viewpoint. We render 420 images for each object.
    \myitem Furniture-Random: We compose 25 dynamic scenes, each containing a random layout of multiple objects and a randomized point light location and viewpoint.
    \myitem Furniture-Realistic: We manually design three scenes containing realistic arrangements of objects in a room with a chosen point light location and viewpoint.
\end{enumerate}
For all three datasets, we use Blender's Cycles path tracer~\cite{blender} to render images at a resolution of 256 by 256 for 
different object arrangements, camera views, and lighting configurations.
For a quantitative evaluation, we report PSNR, SSIM~\cite{wang2003multiscale}, and LPIPS~\cite{zhang2018unreasonable}.

\myparagraph{Baselines and Ablations.}

We compare our method to the following baselines:
\begin{itemize}
    \myitem {\bf o-NeRF}: A variant of the original NeRF model~\cite{mildenhall2020nerf}, but with one NeRF trained per object.  When o-NeRFs are composed into scenes, they are rendered completely separately.
    \myitem {\bf o-NeRF + S}: A extension of o-NeRF with inter-object shadows applied by reducing the incoming light arriving at each o-NeRF by the cumulative opacity of shadowing objects along the ray from the light (Eq.~\ref{eq:tau_m}).
\end{itemize}

\noindent These baselines represent what could be achieved by combining separately trained NeRFs into a scene.   Of course, since o-NeRFs produce radiance fields (not scattering fields), we do not expect them to perform well in novel lighting environments or object placements.  

\begin{figure}
\centering
\includegraphics[width=\linewidth]{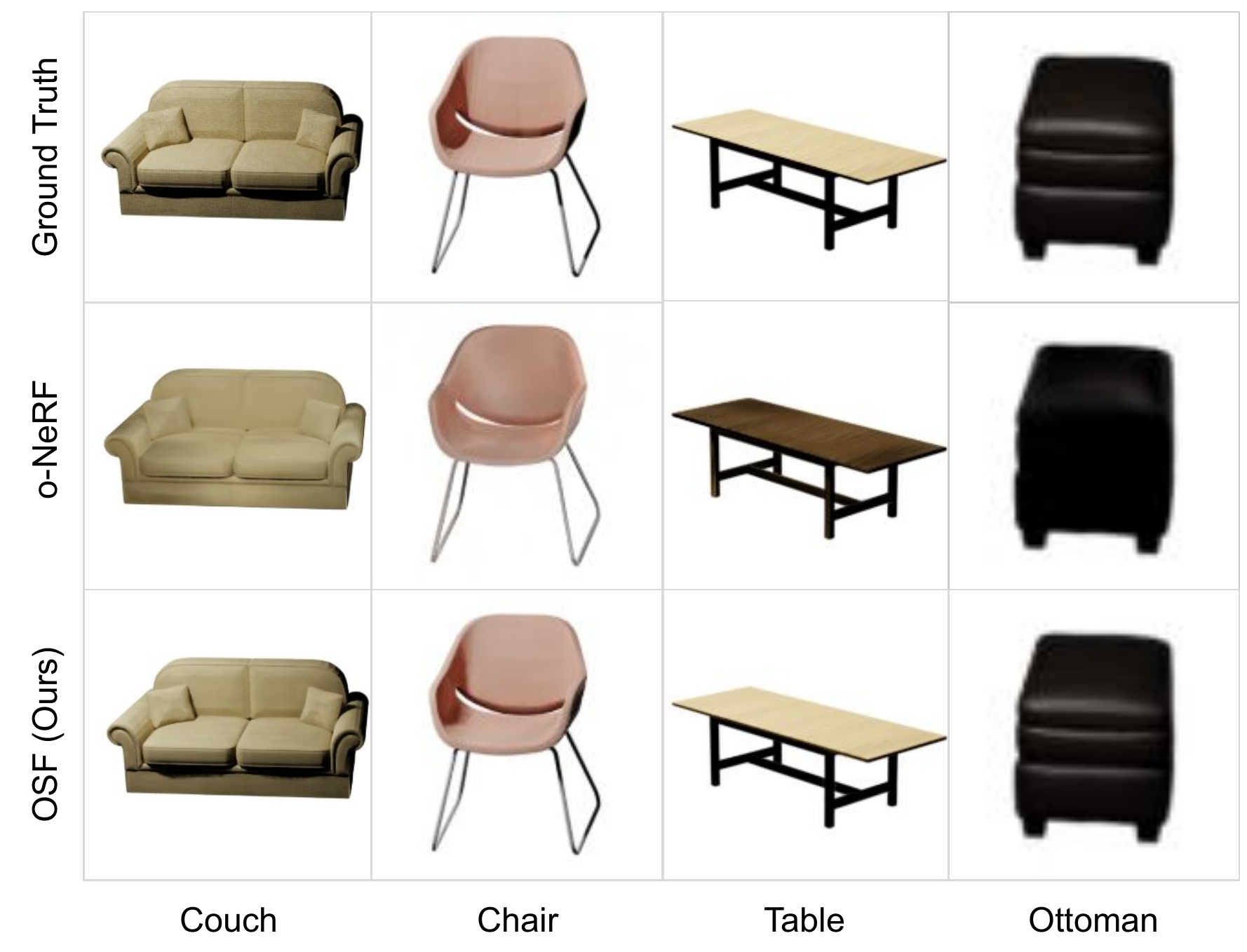}
\caption{Generalization to Novel Illumination. We compare methods on rendering objects under novel lighting directions. Our method (\model) is able to recover fine details in appearance, such as shadows for the couch and the chair, and specular details for the ottoman. o-NeRF fails to disentangle viewpoint versus lighting-dependent appearance, producing incorrect shadows for the couch and chair, and fails to capture the specular details of the ottoman.}
\label{fig:results_single}
\end{figure}

\begin{figure*}[t]
\centering
\includegraphics[width=0.98\linewidth]{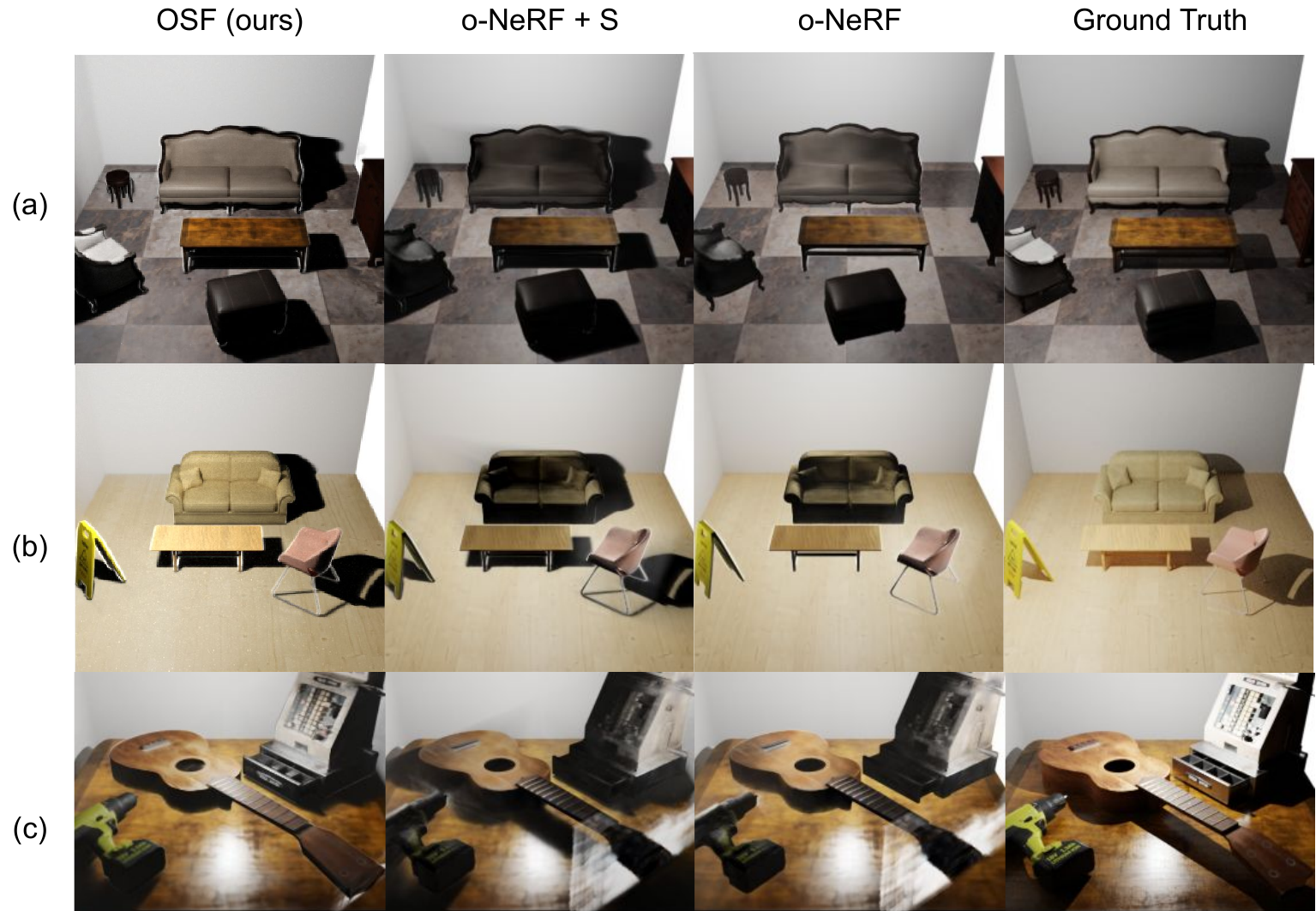}
\caption{Comparisons on scene composition on the Furniture-Realistic dataset.
Columns denote different methods: our model (\model), a variant of NeRF~\cite{mildenhall2020nerf} where one NeRF is trained per object (o-NeRF), and a variant of o-NeRF where shadows are applied (o-NeRF + S).
Compared to o-NeRF, our model (\model) is able to disentangle lighting-dependent appearance from view-dependent appearance for individual objects, and is able to render shadows cast by objects onto the ground correctly.}
\label{fig:furniture_realistic}
\vspace{-10pt}
\end{figure*}

\begin{figure}
\centering
\includegraphics[width=\linewidth]{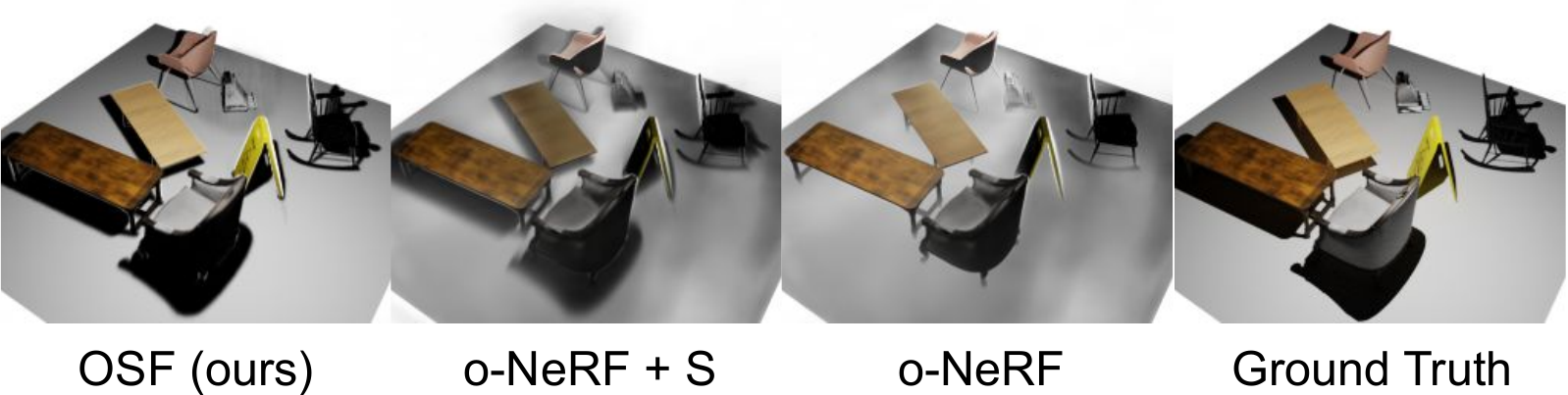}
\caption{Results on scene composition on the Furniture-Random dataset.
We evaluate on our model (\model), a variant of NeRF~\cite{mildenhall2020nerf} where one NeRF is trained per object (o-NeRF), and a version of o-NeRF where shadows are applied (o-NeRF + S).
Compared to o-NeRF, our model (\model) is able to disentangle lighting-dependent and view-dependent appearance and can render shadows.}
\label{fig:furniture_random}
\end{figure}

\subsection{Novel Lighting}
\label{sec:novel_light}
In the first experiment, we investigate how our OSF method handles novel lighting conditions compared to o-NeRF.  
We train one model per object in the Furniture-Single dataset.
For each object model, we train on 400 images with randomized viewpoint and lighting, and test on 20 images of novel viewpoint and lighting. 
Results are shown in \tbl{tbl:single} and \fig{fig:results_single}.
As can be seen in the rendered results, our method produces more accurate appearance of the objects in comparison to o-NeRF when tested on novel illumination conditions.
In particular, o-NeRF fails to predict self-shadows for the couch and chair correctly, and has difficulty recovering the specular details on the ottoman.
This is expected given that o-NeRF assumes fixed illumination and therefore is unable to disentangle view-dependent appearance from lighting-dependent effects.

\begin{table}[t]
\centering
\begin{tabular}{lccc}
\toprule
 \multirow{2}{*}{Methods} & \multicolumn{3}{c}{Furniture-Random} \\ \cmidrule{2-4}
       & \psnr       & \ssim       & \lpips     \\ \midrule
NeRF~\cite{mildenhall2020nerf} & 12.17 & 0.690 & 0.280 \\
o-NeRF + S~\cite{mildenhall2020nerf} & 14.70 & 0.697 & 0.267 \\
\model (Ours) & \textbf{19.02} & \textbf{0.793} & \textbf{0.135} \\
\bottomrule
\end{tabular}
\caption{Results on rendering dynamic scenes with random object poses, lighting, and viewpoints.
Rows denote different methods: our full model (\model), a variant of NeRF~\cite{mildenhall2020nerf} where one NeRF is trained per object (o-NeRF), and a version of o-NeRF where shadows are applied (o-NeRF + S).
}
\label{tbl:furniture_random}
\end{table}

\subsection{Scene Composition}
\label{sec:scene_comp}
In a second experiment, we compare results on a scene composition task, where multiple object models are combined into scenes in random pose, lighting, and viewpoint configurations.
For this task, we use the same object models trained in \sect{sec:novel_light} and evaluate on the Furniture-Random and Furniture-Realistic datasets.
Qualitative results for Furniture-Realistic are shown in \fig{fig:furniture_realistic}.
Results for Furniture-Random appear in \fig{fig:furniture_random} and \tbl{tbl:furniture_random}.
These results suggest that OSF outperforms all baselines and ablations, both quantitatively and qualitatively.
As in the previous experiment (\sect{sec:novel_light}), we find that OSF reproduces object appearances and self-shadows more accurately than the baselines.   The difference is especially apparent in the couches in scenes (a) and (b), where the couches predicted by o-NeRF are extremely dark.
This is due to the fact that o-NeRF is unable to disentangle view-dependence appearance from light-dependent appearance, and simply interpolates the radiance field learned another different lighting configuration.
Please note that \model is able to model inter-object light transport effects by rendering shadows cast by one object onto another and on the ground plane.  Plus, it is able to render indirect illumination of one object reflecting light onto another.  For example, light reflected from the left wall causes the left of the couch and table in scenes (a) and (b) to be brighter.  Neither of these lighting effects are present in the o-NeRF results.

\begin{figure}[t]
\centering
\includegraphics[width=.72\linewidth]{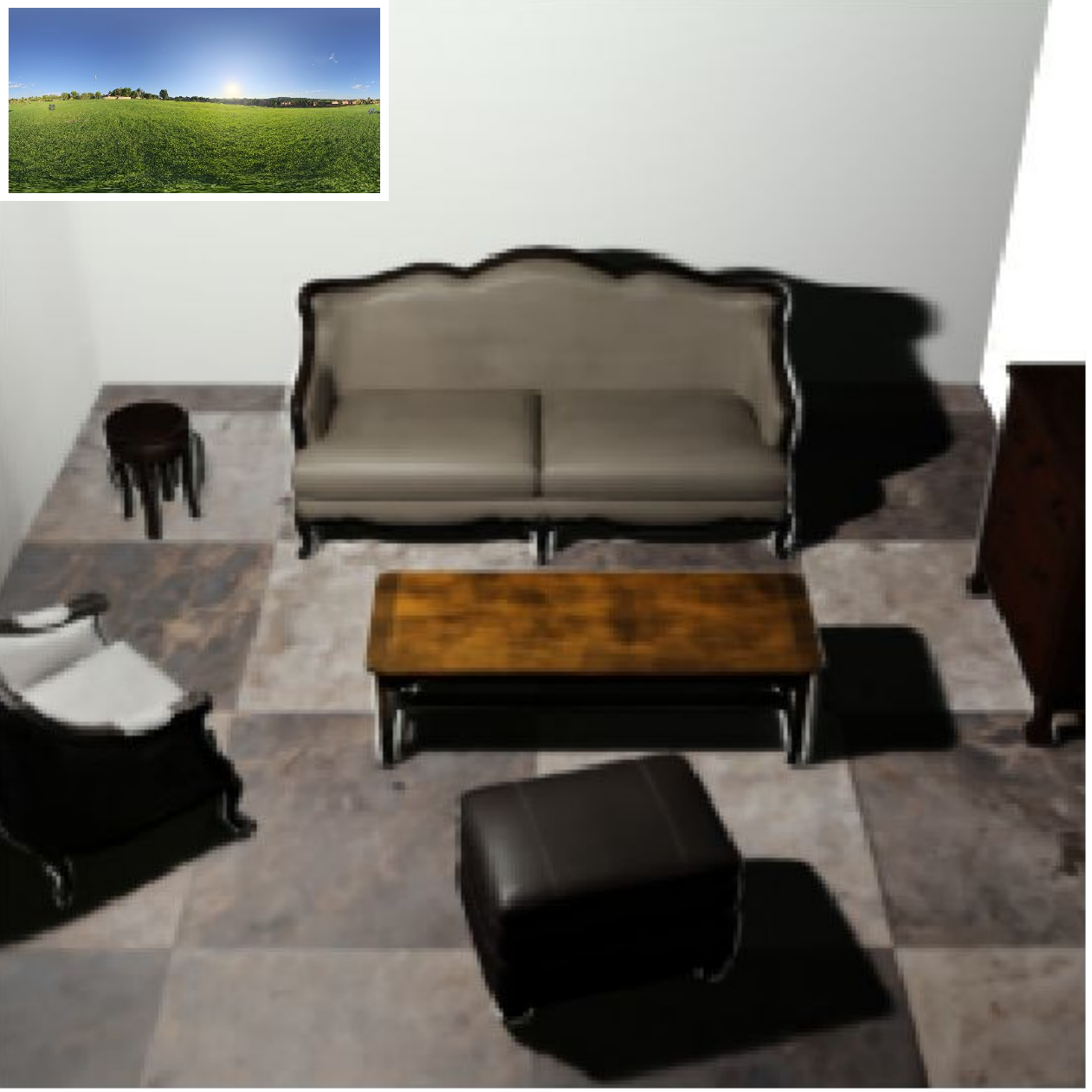}
\caption{Results on rendering with an environment map. We render a scene from Furniture-Realistic with a point light and an outdoor environment map (top-left inset). The environment map causes a blue-green tint in the rendering.}
\label{fig:env_map}
\vspace{-10pt}
\end{figure}

\subsection{Complex Illumination}
\label{sec:env_map}

In a third experiment, we investigate how scenes composed of OSF objects can be rendered with complex illumination from an environment map.
Specifically, we apply the combination of a point light source and the environment map shown in the top-left corner of \fig{fig:env_map} to light one of our scenes in Furniture-Realistic.
This simulates the appearance of the scene as if the scene were inserted into a complex lighting environment, which stresses the benefits of the OSF path tracing framework.  For each OSF sample point, we project the equirectangular coordinates of the environment map into spherical coordinates, sample 20 directions on the unit sphere uniformly at random, evaluate the OSF function for each incoming direction, and integrate them outgoing radiance using Equation~\ref{eq:L_o}.  
The result is shown in \fig{fig:env_map}.   Please note that a green-blue tint is slightly apparent in the scene rendering, due to the contribution of green and blue lighting from the environment map.

\begin{figure}[t]
\centering
\includegraphics[width=\linewidth]{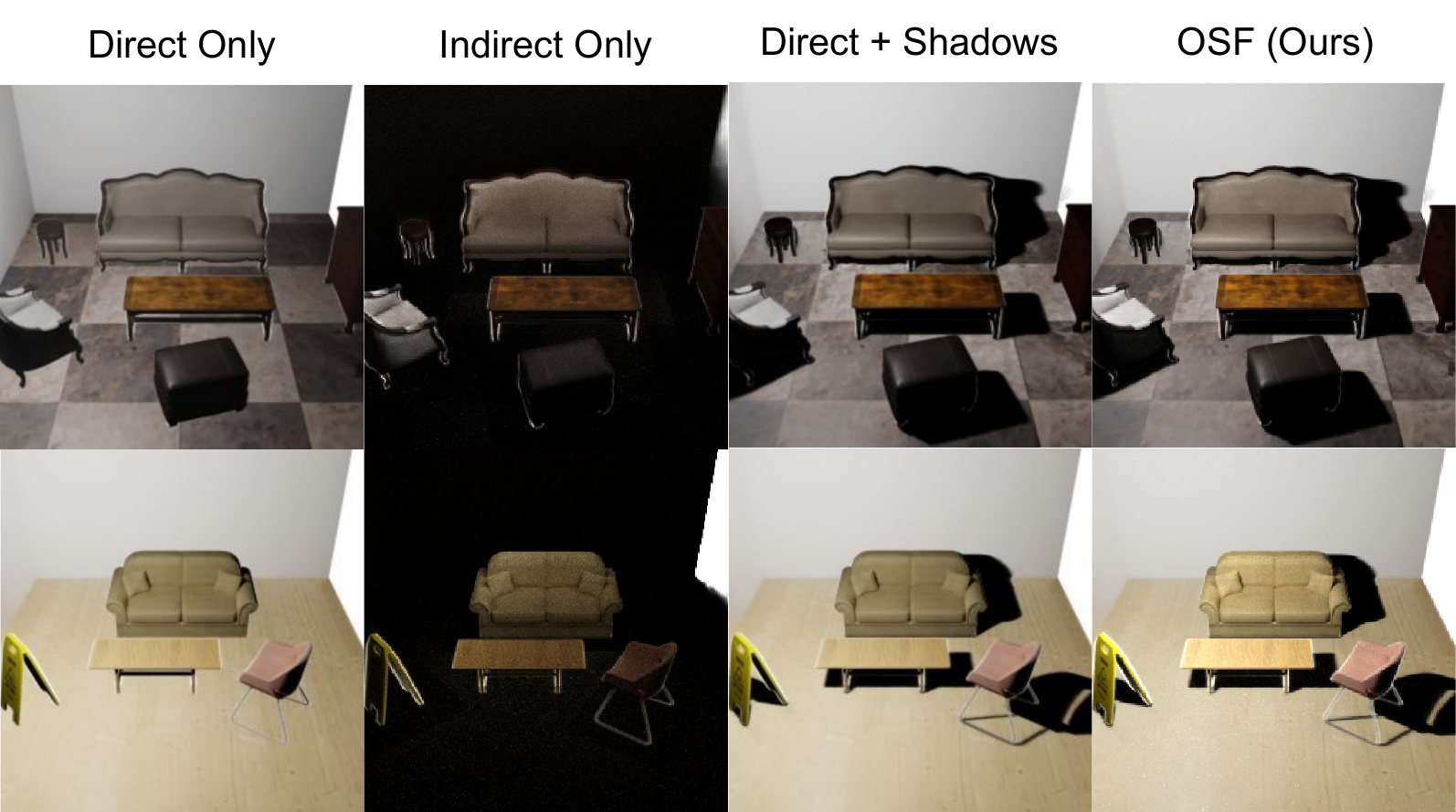}
\caption{Ablation results on our model. We evaluate different variants of our model: ``Direct Only'' which considers only direct illumination; ``Indirect Only'' which considers only indirect illumination; ``Direct + Shadows'' which includes both direct illumination and shadows. Our full model (\model) shows the most realistic rendering, with accurate shadows and indirect illumination effects such as the left side of the couches and tables appearing brighter due to indirect lighting from the left wall.}
\label{fig:ablation}
\vspace{-10pt}
\end{figure}

\subsection{Ablations}
In a final experiment, we perform ablations of our model to study the impact of computing shadows and indirect illumination with our path tracing algorithm (\fig{fig:ablation}).
We show a ``Direct Only" version of our model which represents a version of our model without modeling inter-object lighting effects such as shadows and indirect illumination.
We additionally show ``Direct + Shadows'' and ``Indirect Only'' variants of our model which  represent computing shadows or indirect illumination, respectively.
As can be seen from the figure, our full model containing both shadows and indirect illumination effects is the most realistic.

\section{Conclusion}
We have proposed \models, a method that enables composing objects into photorealistic renderings of dynamic scenes.
We demonstrate that decomposing a scene into implicit object functions that are view- and light-dependent enables reusabiliy of objects across scenes where objects, camera, and lighting can change.
We present a method for integrating our learned implicit functions with volumetric path tracing, and show inter-object light transport effects such as shadow and indirect illumination.  
We believe our work is a step towards a graphics pipeline where real-world scenes are modeled by a composition of implicit functions to combine the flexibility of object-centric neural modeling with the photorealism of graphics rendering algorithms.

\xpar{Acknowledgements.} We thank Konstantinos Rematas for providing the dataset, and Ricardo Martin-Brualla, Pratul Srinivasan, and Jonathon T. Barron for helpful discussions.
MG is funded by a Facebook Fellowship.
We thank 3D Model Haven authors Kirill Sannikov, Joseph Burgan, Ethan Place, Caspian Fortune, Fran Calvente, Joe Seabuhr, and Fernando Quinn for the object models used in our dataset.
We also thank Flaticon.com users DinosoftLabs, Freepik, Good Ware, and Freepik for the icons used in the figure diagrams.

{\small
\bibliographystyle{ieee_fullname}
\balance
\bibliography{egbib}
}

\end{document}